\documentclass[11pt]{article}

\usepackage[final,nohyperref]{acl}

\usepackage{latexsym}
\usepackage[T2A,T1]{fontenc}
\usepackage[utf8]{inputenc}
\usepackage{times}
\usepackage{inconsolata}
\newcommand{\tj}[1]{{\fontencoding{T2A}\fontfamily{cmr}\selectfont #1}}
\DeclareUnicodeCharacter{04E2}{\={\CYRI}}
\DeclareUnicodeCharacter{04E3}{\={\cyri}}
\DeclareUnicodeCharacter{04EE}{\={\CYRU}}
\DeclareUnicodeCharacter{04EF}{\={\cyru}}
\usepackage{microtype}
\usepackage{graphicx}
\usepackage{booktabs}
\usepackage{enumitem}
\setlist{nosep,leftmargin=*}

\newcommand{\BaseLawRten}{80.2}
\newcommand{\BaseLawRone}{52.0}

\newcommand{\BaseLawLo}{77.6}
\newcommand{\BaseLawHi}{82.6}
\newcommand{\BaseTaxRten}{82.6}
\newcommand{\BaseTaxRone}{43.1}

\newcommand{\BaseTaxLo}{75.2}
\newcommand{\BaseTaxHi}{89.9}
\newcommand{\BaseWikiRten}{96.4}
\newcommand{\BaseWikiRone}{89.4}

\newcommand{\BaseWikiLo}{94.8}
\newcommand{\BaseWikiHi}{97.8}
\newcommand{\BaseBelRten}{92.2}
\newcommand{\BaseBelRone}{79.1}

\newcommand{\BaseBelLo}{90.4}
\newcommand{\BaseBelHi}{93.9}
\newcommand{\BaseRuRten}{97.9}

\newcommand{\BaseRuLo}{96.8}
\newcommand{\BaseRuHi}{98.8}
\newcommand{\BaseEnRten}{98.9}

\newcommand{\BaseEnLo}{98.2}
\newcommand{\BaseEnHi}{99.6}

\newcommand{\BaseBelKkRone}{82.4}

\newcommand{\BaseBelUzRone}{86.0}

\newcommand{\BaseBelKyRone}{84.4}

\newcommand{\FtLawRten}{93.6}
\newcommand{\FtLawRone}{68.8}

\newcommand{\FtLawLo}{92.0}
\newcommand{\FtLawHi}{95.1}
\newcommand{\FtTaxRten}{94.5}
\newcommand{\FtTaxRone}{58.7}

\newcommand{\FtTaxLo}{89.9}
\newcommand{\FtTaxHi}{98.2}
\newcommand{\FtWikiRten}{97.4}
\newcommand{\FtWikiRone}{93.2}

\newcommand{\FtWikiLo}{96.0}
\newcommand{\FtWikiHi}{98.6}
\newcommand{\FtBelRten}{95.2}
\newcommand{\FtBelRone}{83.6}

\newcommand{\FtBelLo}{93.9}
\newcommand{\FtBelHi}{96.6}
\newcommand{\FtRuRten}{98.6}

\newcommand{\FtRuLo}{97.8}
\newcommand{\FtRuHi}{99.2}
\newcommand{\FtEnRten}{98.9}

\newcommand{\FtEnLo}{98.1}
\newcommand{\FtEnHi}{99.6}

\newcommand{\FtBelKkRone}{85.4}

\newcommand{\FtBelUzRone}{86.6}

\newcommand{\FtBelKyRone}{85.7}

\newcommand{\FtLatMs}{12}
\newcommand{\CkThreeLawRten}{93.6}

\newcommand{\CkThreeTaxRten}{95.4}

\newcommand{\CkThreeWikiRten}{97.0}

\newcommand{\CkThreeBelRten}{95.7}

\newcommand{\CkFourLawRten}{93.5}

\newcommand{\CkFourTaxRten}{94.5}

\newcommand{\CkFourWikiRten}{97.4}

\newcommand{\CkFourBelRten}{96.1}

\newcommand{\WiseFiveLawRten}{92.1}

\newcommand{\WiseFiveTaxRten}{91.7}

\newcommand{\WiseFiveWikiRten}{96.8}

\newcommand{\WiseFiveBelRten}{95.1}

\newcommand{\WiseSevenLawRten}{93.2}

\newcommand{\WiseSevenTaxRten}{92.7}

\newcommand{\WiseSevenWikiRten}{97.2}

\newcommand{\WiseSevenBelRten}{95.3}

\newcommand{\BaseFoldLawRten}{78.0}

\newcommand{\BaseFoldTaxRten}{78.9}

\newcommand{\BaseFoldWikiRten}{96.2}

\newcommand{\BaseFoldBelRten}{92.0}

\newcommand{\BgeLawRten}{77.6}

\newcommand{\BgeTaxRten}{78.9}

\newcommand{\BgeWikiRten}{95.0}

\newcommand{\BgeBelRten}{83.8}

\newcommand{\BgeRuRten}{98.9}

\newcommand{\BgeEnRten}{99.0}

\newcommand{\BgeFoldLawRten}{76.8}

\newcommand{\BgeFoldTaxRten}{77.1}

\newcommand{\BgeFoldWikiRten}{95.2}

\newcommand{\BgeFoldBelRten}{82.9}

\newcommand{\BgeFtLawRten}{91.6}

\newcommand{\BgeFtTaxRten}{92.7}

\newcommand{\BgeFtWikiRten}{97.2}

\newcommand{\BgeFtBelRten}{92.2}

\newcommand{\BgeFtRuRten}{98.1}

\newcommand{\BgeFtEnRten}{98.1}

\newcommand{\GteLawRten}{62.2}

\newcommand{\GteTaxRten}{66.1}

\newcommand{\GteWikiRten}{92.0}

\newcommand{\GteBelRten}{81.3}

\newcommand{\GteRuRten}{97.3}

\newcommand{\GteEnRten}{98.8}

\newcommand{\GemmaLawRten}{49.3}

\newcommand{\GemmaTaxRten}{45.9}

\newcommand{\GemmaWikiRten}{89.6}

\newcommand{\GemmaBelRten}{69.8}

\newcommand{\GemmaRuRten}{98.6}

\newcommand{\GemmaEnRten}{99.2}

\newcommand{\QwSLawRten}{35.9}

\newcommand{\QwSTaxRten}{51.4}

\newcommand{\QwSWikiRten}{86.6}

\newcommand{\QwSBelRten}{66.1}

\newcommand{\QwSRuRten}{97.7}

\newcommand{\QwSEnRten}{98.9}

\newcommand{\QwMLawRten}{76.4}

\newcommand{\QwMTaxRten}{77.1}

\newcommand{\QwMWikiRten}{95.4}

\newcommand{\QwMBelRten}{88.8}

\newcommand{\QwMRuRten}{99.0}

\newcommand{\QwMEnRten}{99.1}

\newcommand{\QwLLawRten}{76.3}

\newcommand{\QwLLawLo}{73.7}
\newcommand{\QwLLawHi}{78.9}
\newcommand{\QwLTaxRten}{83.5}

\newcommand{\QwLWikiRten}{96.4}

\newcommand{\QwLBelRten}{96.8}

\newcommand{\QwLBelLo}{95.7}
\newcommand{\QwLBelHi}{97.9}
\newcommand{\QwLRuRten}{99.7}

\newcommand{\QwLEnRten}{99.8}

\newcommand{\FtwoLawRten}{69.7}

\newcommand{\FtwoTaxRten}{69.7}

\newcommand{\FtwoWikiRten}{94.8}

\newcommand{\FtwoBelRten}{94.0}

\newcommand{\FtwoRuRten}{98.3}

\newcommand{\FtwoEnRten}{99.0}

\newcommand{\GigaLawRten}{92.2}

\newcommand{\GigaLawLo}{90.5}
\newcommand{\GigaLawHi}{93.8}
\newcommand{\GigaTaxRten}{89.9}

\newcommand{\GigaWikiRten}{98.6}

\newcommand{\GigaBelRten}{97.1}

\newcommand{\GigaBelLo}{96.0}
\newcommand{\GigaBelHi}{98.2}
\newcommand{\GigaRuRten}{98.8}

\newcommand{\GigaEnRten}{98.9}

\newcommand{\GigaLatMs}{35}
\newcommand{\GigaFtLawRten}{94.6}

\newcommand{\GigaFtTaxRten}{96.3}

\newcommand{\GigaFtWikiRten}{99.0}

\newcommand{\GigaFtBelRten}{97.8}

\newcommand{\GigaFtBelLo}{96.8}
\newcommand{\GigaFtBelHi}{98.7}
\newcommand{\GigaFtRuRten}{99.3}

\newcommand{\GigaFtEnRten}{99.2}

\newcommand{\QwLFtLawRten}{92.8}

\newcommand{\QwLFtTaxRten}{92.7}

\newcommand{\QwLFtWikiRten}{97.2}

\newcommand{\QwLFtBelRten}{98.2}

\newcommand{\QwLFtBelLo}{97.3}
\newcommand{\QwLFtBelHi}{99.1}
\newcommand{\QwLFtRuRten}{99.7}

\newcommand{\QwLFtEnRten}{99.8}

\newcommand{\QwLFtLatMs}{38}
\newcommand{\FtDeltaLawRten}{+13.4}
\newcommand{\FtDeltaLawRone}{+16.8}
\newcommand{\BgeFtDeltaLawRten}{+14.0}

\newcommand{\FtDeltaTaxRten}{+11.9}
\newcommand{\FtDeltaTaxRone}{+15.6}
\newcommand{\BgeFtDeltaTaxRten}{+13.8}

\newcommand{\FtDeltaWikiRten}{+1.0}
\newcommand{\FtDeltaWikiRone}{+3.8}

\newcommand{\FtDeltaBelRten}{+3.0}
\newcommand{\FtDeltaBelRone}{+4.4}
\newcommand{\BgeFtDeltaBelRten}{+8.4}

\newcommand{\FtDeltaRuRten}{+0.7}

\newcommand{\FtDeltaEnRten}{+0.0}

\newcommand{\BaseFourMean}{87.8}
\newcommand{\BaseFoldFourMean}{86.3}
\newcommand{\BgeFourMean}{83.8}
\newcommand{\BgeFoldFourMean}{83.0}
\newcommand{\UnkLetters}{\tj{ҷ Ӣ Ӯ Ҷ}}
\newcommand{\UnkCount}{4}
\newcommand{\LetterCount}{12}

\newcommand{\NLawQ}{1,000}
\newcommand{\NLawCorpus}{72,920}
\newcommand{\NTaxQ}{109}
\newcommand{\NTaxCorpus}{23,282}
\newcommand{\NWikiQ}{500}
\newcommand{\NWikiCorpus}{15,351}
\newcommand{\NBelQ}{900}
\newcommand{\NBelCorpus}{488}
\newcommand{\NRuQ}{900}

\newcommand{\NEnQ}{900}

\newcommand{\NLawHeld}{1,978}
\newcommand{\NLawAll}{19,895}
\newcommand{\NTaxHeld}{22}
\newcommand{\NTaxAll}{233}
\newcommand{\NWikiHeld}{750}
\newcommand{\NPairsTotal}{235,140}
\newcommand{\NPairsLaw}{52,758}
\newcommand{\NPairsLawNeg}{40,259}
\newcommand{\NPairsTax}{1,033}
\newcommand{\NPairsTaxNeg}{875}
\newcommand{\NPairsWiki}{21,123}
\newcommand{\NPairsWikiNeg}{20,374}
\newcommand{\NPairsWeb}{59,985}
\newcommand{\NPairsWebNeg}{39,140}
\newcommand{\NPairsSvc}{15,744}
\newcommand{\NPairsSvcNeg}{6,313}
\newcommand{\NPairsEn}{49,987}
\newcommand{\NPairsRuMiracl}{4,664}
\newcommand{\NPairsRuHnp}{29,846}

\newcommand{\NPairsWithNeg}{106,961}
\newcommand{\PctReplay}{36}
\newcommand{\PctWithNeg}{71}
\newcommand{\LoraPct}{40}
\newcommand{\SizeRatioLo}{5}
\newcommand{\SizeRatioHi}{14}
\newcommand{\TrainSteps}{466}
\newcommand{\TrainLr}{$2\times10^{-5}$}
\newcommand{\TrainMinutes}{56}
\newcommand{\LossFirst}{4.60}
\newcommand{\LossLastMean}{0.43}
\newcommand{\BfZeroStep}{8}
\newcommand{\NLayers}{24}
\newcommand{\Hidden}{1,024}
\newcommand{\Vocab}{250,002}

\newcommand{\ModelName}{Saidzoda-Tajik-E5-Large}

\title{\ModelName: Fine-Tuning a 560M Embedder\\for Tajik Retrieval}

\author{Tohir Saidzoda \and Nekruz Maxudov \and Bedil Karimov \\
  \textbf{Abubakr Abdullozoda} \and \textbf{Amin Jalilov} \and \textbf{Jonibek Kishvarov} \\
  \textbf{Rahmonali Sharipov} \and \textbf{Mariyamkhon Shodyeva} \\
  Saidzoda AI Lab, Saidzoda Engineering Company, Dushanbe, Tajikistan \\
  \texttt{tohir@saidzoda.com}}

\begin{document}
\maketitle

\begin{abstract}
Finding the passage that answers a question is the first step of most practical systems built on documents, and for Tajik we found no openly described model for it.
We present \ModelName, a Tajik text embedder built from multilingual-e5-large-instruct (560M parameters).
We asked our own Tajik language model to write questions for passages from Tajik laws, tax documents, Wikipedia and web text, and trained on \NPairsTotal{} question and passage pairs for one epoch, with hard negatives chosen by another embedder and English and Russian pairs mixed in so the model would not forget those languages.
On laws the model had never seen, the right passage is among the top ten results for \FtLawRten\% of questions, up from \BaseLawRten\%; on tax documents, \FtTaxRten\% against \BaseTaxRten\%.
On Belebele, the one test we have that was written by people, the gain is smaller and not confirmed (\BaseBelRten\% to \FtBelRten\%, with touching intervals), and Russian and English do not change.
On law and tax questions written in the same style as its training questions, it is tied with Giga-Embeddings-3B, a model five times its size; on Belebele the 3B and 8B models score slightly higher, and only after we fine-tuned them on our data are they clearly ahead.
We also report what failed: training in bfloat16 broke the model, and replacing Tajik letters with similar Russian ones did not help.
Our law, tax and Wikipedia test questions come from the same model that wrote the training questions, so we treat those scores as an upper limit until questions written by Tajik speakers are available.

\end{abstract}

\section{Introduction}

Tajik, the state language of Tajikistan, is written in a Cyrillic alphabet with six letters that Russian does not have (\tj{ғ, ӣ, қ, ӯ, ҳ, ҷ}).
When a citizen asks which article of a law applies to them, or a small business owner asks how a tax is calculated, a useful system first has to find the right passage among tens of thousands.
That step is done by a text embedder, a model that places a question close to the passages that answer it.

Today's multilingual embedders treat Tajik as an afterthought.
They learn mostly from high-resource languages, and the tokenizer shared by multilingual E5 \citep{wang2024me5}, BGE-M3 \citep{chen2024bgem3} and other models built on XLM-RoBERTa \citep{conneau2020xlmr} does not even know \UnkCount{} of the \LetterCount{} Tajik letters (upper and lower case): \UnkLetters{} all become the same unknown token.
We found no published work on Tajik retrieval or Tajik embedders.
The nearest work is on Persian \citep{zinvandi2025famteb,sarmadi2025hakim}, which is written in a different script.

What has changed is that Tajik now has open language models \citep{liashkov2026soro} and large web corpora \citep{arabov2026tajikweb}, and a language model that writes Tajik well can produce training questions for any passage.
We used our own Tajik language model for this, and in this report we describe the embedder we trained with its questions, how we tested it, and where it still falls short.

Our contributions are:
\begin{enumerate}
\item \textbf{The model and how we made it.} A 560M embedder fine-tuned for Tajik in about \TrainMinutes{} minutes on one GPU, with every step of the recipe: how the training pairs were made, how negatives were chosen, and the settings we used (Sections~\ref{sec:data} and~\ref{sec:recipe}).
\item \textbf{A Tajik retrieval test.} Questions on laws, tax documents and Wikipedia articles that were kept out of training as whole units, scored together with the Tajik part of Belebele retrieval \citep{bandarkar2024belebele,enevoldsen2025mmteb} (Section~\ref{sec:data}).
\item \textbf{A fair comparison.} Nine embedders from 0.3B to 8B parameters used as released, and three more fine-tuned on our pairs, all scored on the same questions (Section~\ref{sec:results}).
\item \textbf{What did not work.} Training in bfloat16, replacing Tajik letters with Russian ones, and mixing the trained weights back with the original weights (Section~\ref{sec:negative}).
\end{enumerate}

\section{Related Work}

\paragraph{Multilingual text embedders.}
Sentence-BERT \citep{reimers2019sbert} established fine-tuning a pretrained encoder with a contrastive objective for retrieval.
E5 \citep{wang2022e5} and its multilingual successor \citep{wang2024me5} add large-scale weakly supervised pretraining and, in the instruct variant, a task instruction on the query side.
BGE-M3 \citep{chen2024bgem3} and mGTE \citep{zhang2024mgte} are other XLM-RoBERTa-based multilingual embedders.
A newer group of embedders is built on decoder language models: Qwen3-Embedding \citep{zhang2025qwen3emb}, F2LLM \citep{zhang2025f2llm,zhang2026f2llmv2} and the Giga-Embeddings family \citep{kolodin2025gigaembeddings}; EmbeddingGemma \citep{vera2025embeddinggemma} is a small model of this kind.
We found no report of Tajik retrieval results for these models beyond the Belebele task in MMTEB. MTEB and MMTEB \citep{muennighoff2023mteb,enevoldsen2025mmteb} are the standard benchmarks; for Tajik, MMTEB includes Belebele used as retrieval, which we also use, but no domain retrieval task such as law or tax.

\paragraph{Contrastive training details.}
We use the in-batch multiple-negatives loss, building on \citet{henderson2017smartreply}, with the gradient caching of \citet{gao2021gradcache} so that a batch of 512 fits on one GPU.
Hard negatives are mined with an existing embedder and filtered with the positive-aware margin of NV-Retriever \citep{moreira2024nvretriever} to avoid false negatives.
WiSE-FT \citep{wortsman2022wiseft} interpolates fine-tuned and original weights to keep general ability; we test it as an ablation.

\paragraph{Synthetic queries.}
Training retrievers on queries written by a language model for existing passages goes back to InPars, Promptagator and GPL \citep{bonifacio2022inpars,dai2022promptagator,wang2022gpl}, and has been applied to many languages \citep{thakur2024swimir}.
Whether test collections built with LLMs favour systems close to the generator is debated: \citet{rahmani2024synthetic} found little such bias, while \citet{rahmani2025bias} report that it distorts absolute scores more than system rankings.
Our law, tax and Wikipedia test queries come from the same generator as our training queries, so we read their absolute values as upper bounds and rely on Belebele for an independent check.

\paragraph{Tajik and nearby languages.}
Tajik language technology has recently gained web corpora \citep{arabov2026tajikweb}, NLP resources \citep{arabov2026tajiknlp} and large language models \citep{liashkov2026soro}, but we found no Tajik retrieval benchmark or embedder.
Persian, the closest major language, has a benchmark \citep{zinvandi2025famteb} and dedicated embedders \citep{sarmadi2025hakim}; it is written in the Perso-Arabic script, so these do not transfer directly.

\section{Data and Benchmark}
\label{sec:data}

\paragraph{Sources.}
All Tajik passages come from four collections: a corpus of Tajik laws split into \NLawCorpus{} chunks, a collection of tax-code and tax-guidance documents split into \NTaxCorpus{} chunks, Tajik Wikipedia, and cleaned Tajik web and literature text.
For each passage our Tajik fine-tune of Qwen3.6-27B wrote questions that the passage answers, and a second, independent call of the model checked each question against its passage; only checked questions were kept.
We use these question and passage pairs both for training and, on held-out material, for testing.

\paragraph{Held-out split.}
Before any training pair was built, we froze the test material at the level of whole units, so that no test passage and no other passage from the same unit is seen in training or used as a negative.
For law, \NLawHeld{} of \NLawAll{} laws are held out; for tax, \NTaxHeld{} of \NTaxAll{} documents; for Wikipedia, \NWikiHeld{} articles.
The choice is made by the SHA-1 hash of the unit's title (about one law or document in ten and one article in twenty), and the lists are stored with the benchmark; the test corpora also contain the non-held-out passages as distractors.
In addition, any training query whose normalised text equals a test query is dropped.

\paragraph{Test sets.}
\begin{itemize}
\item \textbf{Law}: \NLawQ{} questions on held-out laws; the corpus is all \NLawCorpus{} law chunks, so the model must find the right chunk among every law, not only the held-out ones.
\item \textbf{Tax}: \NTaxQ{} questions on held-out documents, over all \NTaxCorpus{} tax chunks. This set is small, and its intervals are wide.
\item \textbf{Wikipedia}: \NWikiQ{} questions on held-out articles, over \NWikiCorpus{} Wikipedia passages.
\item \textbf{Belebele} \citep{bandarkar2024belebele}: the \NBelQ{} human-written Tajik questions, each to be matched with its own passage among the \NBelCorpus{} distinct passages, as in the Belebele retrieval task of MMTEB \citep{enevoldsen2025mmteb}, except that we report recall rather than nDCG. This is the only test set not produced by our generator. The Russian and English versions of the same set (\NRuQ{} questions each) check that the other languages are not lost.
\end{itemize}
The main metric is recall@10, the share of questions whose gold passage is among the ten nearest passages; we also report recall@1.
Each recall comes with a 95\% bootstrap interval over questions (1,000 resamples); the passages are held fixed, which matters for Belebele, where several questions share a passage.
All models are scored with the same script (sentence-transformers 5.7), in bfloat16, with each model's own query instruction, and with its document instruction for passages where the model defines one.

\paragraph{Training pairs.}
Table~\ref{tab:data} lists the training pairs by domain.
The Tajik pairs come from the same generator on non-held-out material.
A further \NPairsSvc{} pairs teach the model to map short customer-service requests to topic descriptions; the requests were written by a language model in a simulation of our own customer-service line, and this set is not released.
To keep the other languages, \PctReplay\% of the pairs are replay: English Natural Questions \citep{kwiatkowski2019nq}, Russian MIRACL \citep{zhang2023miracl} and the Russian hard-negative set ru-HNP.
The Russian MIRACL replay is smaller than planned (\NPairsRuMiracl{} pairs) because rows with a repeated query are removed when the pairs are built.

\begin{table}[t]\centering\footnotesize\setlength{\tabcolsep}{4pt}
\begin{tabular}{lrr}
\toprule
Domain & Pairs & With negative \\
\midrule
Tajik law & \NPairsLaw & \NPairsLawNeg \\
Tajik tax & \NPairsTax & \NPairsTaxNeg \\
Tajik Wikipedia & \NPairsWiki & \NPairsWikiNeg \\
Tajik web and literature & \NPairsWeb & \NPairsWebNeg \\
Customer service (synthetic) & \NPairsSvc & \NPairsSvcNeg \\
\midrule
English replay (NQ) & \NPairsEn & n/a \\
Russian replay (MIRACL) & \NPairsRuMiracl & n/a \\
Russian replay (HNP set) & \NPairsRuHnp & n/a \\
\midrule
Total & \NPairsTotal & \NPairsWithNeg \\
\bottomrule
\end{tabular}
\caption{Training pairs by domain, as counted by the training run. Replay pairs use in-batch negatives only.}
\label{tab:data}
\end{table}

\section{Training Recipe}
\label{sec:recipe}

\paragraph{Starting model.}
We start from multilingual-e5-large-instruct \citep{wang2024me5}, an XLM-RoBERTa-large encoder (\NLayers{} layers, hidden size \Hidden, vocabulary \Vocab).
We chose it in a first round in which it had the best mean recall among five small embedders (Table~\ref{tab:compare} shows all of them); that round used the same law and tax test sets reported here, plus an in-house customer-service set.
The query instruction of the original model (``Given a web search query, retrieve relevant passages that answer the query'') is kept on the query side in training and at use time.

\paragraph{Hard negatives.}
For every in-domain pair we searched the passage pool of its domain with BGE-M3 and took one negative at random from ranks 10 to 50, under three conditions: it comes from a different law, document or article than the positive, it is not the positive text, and its score is below 0.95 times the score of the positive \citep{moreira2024nvretriever}.
\NPairsWithNeg{} pairs (\PctWithNeg\% of the in-domain pairs) received a negative; for the rest no candidate met all three conditions, and they train with in-batch negatives only, as do all replay pairs.

\begin{figure}[t]\centering
\includegraphics[width=0.9\columnwidth]{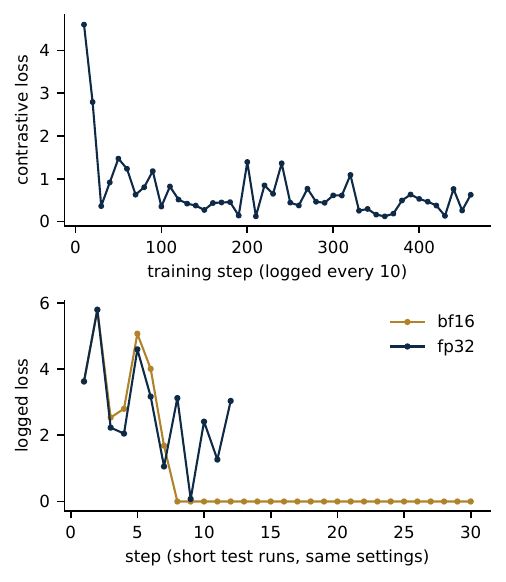}
\caption{Top: training loss of \ModelName{} (logged every 10 steps). Bottom: two short test runs with the same settings; in bfloat16 the logged loss drops to exactly zero at step \BfZeroStep{} and stays there, in fp32 it does not.}
\label{fig:loss}
\end{figure}

\paragraph{Loss and settings.}
We train with the cached multiple-negatives ranking loss of sentence-transformers (in-batch negatives plus the mined negative, gradient caching with mini-batches of 64, scale 20), with a global batch of 512 on one H100 GPU.
Each batch is drawn from a single domain, so all in-batch negatives are in the same domain as the positive; domains are sampled in proportion to their size, and a batch never contains the same text twice.
The learning rate is \TrainLr{} with a linear schedule and 5\% warm-up, for one epoch (\TrainSteps{} steps, slightly more than the number of pairs divided by 512 because batches are filled per domain; about \TrainMinutes{} minutes), with sequences up to 512 tokens and seed 42.
Training runs in fp32 with TF32 matrix multiplication; bfloat16 failed (Section~\ref{sec:negative}).
The training loss falls from \LossFirst{} at step 10 to a mean of \LossLastMean{} over the last five logged steps (Figure~\ref{fig:loss}).

\section{Results}
\label{sec:results}

\begin{table*}[t]\centering\small
\begin{tabular}{lrrr}
\toprule
Test set (queries) & e5 base & Ours & Change \\
\midrule
Law (\NLawQ) & \BaseLawRten\ {\scriptsize[\BaseLawLo, \BaseLawHi]} & \FtLawRten\ {\scriptsize[\FtLawLo, \FtLawHi]} & \FtDeltaLawRten \\
Tax (\NTaxQ) & \BaseTaxRten\ {\scriptsize[\BaseTaxLo, \BaseTaxHi]} & \FtTaxRten\ {\scriptsize[\FtTaxLo, \FtTaxHi]} & \FtDeltaTaxRten \\
Wikipedia (\NWikiQ) & \BaseWikiRten\ {\scriptsize[\BaseWikiLo, \BaseWikiHi]} & \FtWikiRten\ {\scriptsize[\FtWikiLo, \FtWikiHi]} & \FtDeltaWikiRten \\
Belebele Tajik (\NBelQ) & \BaseBelRten\ {\scriptsize[\BaseBelLo, \BaseBelHi]} & \FtBelRten\ {\scriptsize[\FtBelLo, \FtBelHi]} & \FtDeltaBelRten \\
Belebele Russian (\NRuQ) & \BaseRuRten\ {\scriptsize[\BaseRuLo, \BaseRuHi]} & \FtRuRten\ {\scriptsize[\FtRuLo, \FtRuHi]} & \FtDeltaRuRten \\
Belebele English (\NEnQ) & \BaseEnRten\ {\scriptsize[\BaseEnLo, \BaseEnHi]} & \FtEnRten\ {\scriptsize[\FtEnLo, \FtEnHi]} & \FtDeltaEnRten \\
\midrule
\multicolumn{4}{l}{\emph{Recall@1}} \\
Law & \BaseLawRone & \FtLawRone & \FtDeltaLawRone \\
Tax & \BaseTaxRone & \FtTaxRone & \FtDeltaTaxRone \\
Wikipedia & \BaseWikiRone & \FtWikiRone & \FtDeltaWikiRone \\
Belebele Tajik & \BaseBelRone & \FtBelRone & \FtDeltaBelRone \\
\bottomrule
\end{tabular}
\caption{Recall@10 (top block, with 95\% bootstrap intervals) and recall@1 (bottom block) in percent, for the starting model and our fine-tuned model. Change is in percentage points. Law, tax and Wikipedia queries were written by an LLM on held-out documents; only the Belebele sets are human-written.}
\label{tab:main}
\end{table*}

\paragraph{Gains over the starting model.}
Table~\ref{tab:main} compares \ModelName{} with the model it was trained from.
The clearest gain is on law: recall@10 rises from \BaseLawRten{} to \FtLawRten{} (\FtDeltaLawRten{} points), and the 95\% intervals, [\BaseLawLo, \BaseLawHi] and [\FtLawLo, \FtLawHi], do not overlap.
Tax rises by \FtDeltaTaxRten{} points; with only \NTaxQ{} questions the intervals meet at their ends, so we read this as a likely but not certain gain.
Wikipedia, which the starting model already handles well, changes by \FtDeltaWikiRten{} point, within the noise.
We repeat that these three sets were produced by the same generator as the training queries, so their absolute values are optimistic.

On Belebele Tajik, which is human-written and was not used to build any training pair, recall@10 rises by \FtDeltaBelRten{} points and recall@1 by \FtDeltaBelRone{} points.
The recall@10 intervals touch ([\BaseBelLo, \BaseBelHi] and [\FtBelLo, \FtBelHi]), and we did not store per-question ranks, so we cannot run a paired test.
As with tax, we read this as a likely small gain rather than a confirmed one, and it is much smaller than on our own sets.
Russian and English are unchanged (\BaseRuRten{} to \FtRuRten{} and \BaseEnRten{} to \FtEnRten). We did not train without replay, so we cannot say how much of this is due to the replay pairs.

The same pairs and settings applied to BGE-M3 give similar gains on law (\BgeFtDeltaLawRten) and tax (\BgeFtDeltaTaxRten) and a larger one on Belebele Tajik (\BgeFtDeltaBelRten{} points, with intervals that do not overlap), while Russian and English fall by less than one point (Table~\ref{tab:compare}).
The recipe is therefore not tied to one starting model; BGE-M3 starts lower than E5 and ends slightly lower, though not by more than the intervals on law and tax.

\begin{table*}[t]\centering\small
\begin{tabular}{llrrrrrrr}
\toprule
Model & Size & Used as & Law & Tax & Wiki & Bel.\ tg & Bel.\ ru & Bel.\ en \\
\midrule
EmbeddingGemma-300M & 0.31B & zero-shot & \GemmaLawRten & \GemmaTaxRten & \GemmaWikiRten & \GemmaBelRten & \GemmaRuRten & \GemmaEnRten \\
gte-multilingual-base & 0.31B & zero-shot & \GteLawRten & \GteTaxRten & \GteWikiRten & \GteBelRten & \GteRuRten & \GteEnRten \\
Qwen3-Embedding-0.6B & 0.6B & zero-shot & \QwSLawRten & \QwSTaxRten & \QwSWikiRten & \QwSBelRten & \QwSRuRten & \QwSEnRten \\
BGE-M3 & 0.57B & zero-shot & \BgeLawRten & \BgeTaxRten & \BgeWikiRten & \BgeBelRten & \BgeRuRten & \BgeEnRten \\
multilingual-e5-large-instruct & 0.56B & zero-shot & \BaseLawRten & \BaseTaxRten & \BaseWikiRten & \BaseBelRten & \BaseRuRten & \BaseEnRten \\
F2LLM-v2-4B & 4B & zero-shot & \FtwoLawRten & \FtwoTaxRten & \FtwoWikiRten & \FtwoBelRten & \FtwoRuRten & \FtwoEnRten \\
Qwen3-Embedding-4B & 4B & zero-shot & \QwMLawRten & \QwMTaxRten & \QwMWikiRten & \QwMBelRten & \QwMRuRten & \QwMEnRten \\
Giga-Embeddings-instruct-3B & 3B & zero-shot & \GigaLawRten & \GigaTaxRten & \GigaWikiRten & \GigaBelRten & \GigaRuRten & \GigaEnRten \\
Qwen3-Embedding-8B & 8B & zero-shot & \QwLLawRten & \QwLTaxRten & \QwLWikiRten & \QwLBelRten & \QwLRuRten & \QwLEnRten \\
\midrule
BGE-M3, same recipe & 0.57B & full FT & \BgeFtLawRten & \BgeFtTaxRten & \BgeFtWikiRten & \BgeFtBelRten & \BgeFtRuRten & \BgeFtEnRten \\
Giga-Embeddings-3B, LoRA & 3B & LoRA, \LoraPct\% pairs & \textbf{\GigaFtLawRten} & \textbf{\GigaFtTaxRten} & \textbf{\GigaFtWikiRten} & \GigaFtBelRten & \GigaFtRuRten & \GigaFtEnRten \\
Qwen3-Embedding-8B, LoRA & 8B & LoRA, \LoraPct\% pairs & \QwLFtLawRten & \QwLFtTaxRten & \QwLFtWikiRten & \textbf{\QwLFtBelRten} & \textbf{\QwLFtRuRten} & \textbf{\QwLFtEnRten} \\
Ours (e5, fine-tuned) & 0.56B & full FT & \FtLawRten & \FtTaxRten & \FtWikiRten & \FtBelRten & \FtRuRten & \FtEnRten \\
\bottomrule
\end{tabular}
\caption{Recall@10 (percent) of every embedder we scored on the same frozen test sets. Bold marks the highest value in each column, not a significant difference. Rows above the line are used as released; rows below were trained on our pairs (the LoRA runs used \LoraPct\% of them). Belebele Russian and English check that the other languages are not lost.}
\label{tab:compare}
\end{table*}

\paragraph{Comparison with other embedders.}
Table~\ref{tab:compare} places the model among nine embedders used as released and three more trained on our pairs.
Used as released, the small embedders are weak on Tajik law: EmbeddingGemma-300M reaches \GemmaLawRten{} and Qwen3-Embedding-0.6B \QwSLawRten, even though their tokenizers cover every Tajik letter.
Size alone does not fix this: Qwen3-Embedding-8B reaches \QwLLawRten{} on law [\QwLLawLo, \QwLLawHi], well below our \FtLawRten.
The strongest model used as released is Giga-Embeddings-instruct-3B\footnote{\texttt{ai-sage/Giga-Embeddings-instruct-3B-0826} on the Hugging Face Hub.}: on law (\GigaLawRten, [\GigaLawLo, \GigaLawHi]) and tax (\GigaTaxRten) its intervals overlap ours, so the two are tied there.
On Belebele Tajik it scores higher than ours (\GigaBelRten{} against \FtBelRten), as does Qwen3-Embedding-8B (\QwLBelRten), but the intervals overlap ([\GigaBelLo, \GigaBelHi] and [\QwLBelLo, \QwLBelHi] against [\FtBelLo, \FtBelHi]), so we cannot call this a difference.
The comparison also favours our model on law and tax, because its training questions share the style of the test questions and the other models' did not.

When the two large models are fine-tuned on our pairs with LoRA (on \LoraPct\% of the pairs, because of their training cost), Giga-Embeddings-3B gives the best law, tax and Wikipedia scores in the table and Qwen3-Embedding-8B the best Belebele score.
This comparison is like for like, since all three models saw the same kind of questions. On law and tax their intervals overlap ours; on Belebele both are clearly ahead ([\GigaFtBelLo, \GigaFtBelHi] and [\QwLFtBelLo, \QwLFtBelHi]).
In short, the tuned 560M model is as good as models \SizeRatioLo{} to \SizeRatioHi{} times larger on the in-domain Tajik tasks, and a larger model trained the same way is better at the human-written reading questions.
At one query at a time, \ModelName{} needed about \FtLatMs{} ms per query on an H100 and the 3B and 8B models about \GigaLatMs{} to \QwLFtLatMs{} ms; these timings were taken on a shared machine and vary, so we give them only as a rough guide.

\paragraph{Other Central Asian languages.}
Although we trained on Tajik only, recall@1 on Belebele Kazakh, Uzbek and Kyrgyz also moved slightly upward (Kazakh \BaseBelKkRone{} to \FtBelKkRone, Uzbek \BaseBelUzRone{} to \FtBelUzRone, Kyrgyz \BaseBelKyRone{} to \FtBelKyRone).
We note this as an observation and make no claim of transfer; we did not test whether these changes are larger than the noise.

\section{What Did Not Work}
\label{sec:negative}

\begin{table}[t]\centering\small
\begin{tabular}{lrrrr}
\toprule
Variant & Law & Tax & Wiki & Bel.\ tg \\
\midrule
e5 base & \BaseLawRten & \BaseTaxRten & \BaseWikiRten & \BaseBelRten \\
e5 base, letters folded & \BaseFoldLawRten & \BaseFoldTaxRten & \BaseFoldWikiRten & \BaseFoldBelRten \\
\midrule
Ours, step 300 & \CkThreeLawRten & \CkThreeTaxRten & \CkThreeWikiRten & \CkThreeBelRten \\
Ours, step 400 & \CkFourLawRten & \CkFourTaxRten & \CkFourWikiRten & \CkFourBelRten \\
Ours, final (step \TrainSteps) & \FtLawRten & \FtTaxRten & \FtWikiRten & \FtBelRten \\
WiSE, 0.7 tuned + 0.3 base & \WiseSevenLawRten & \WiseSevenTaxRten & \WiseSevenWikiRten & \WiseSevenBelRten \\
WiSE, 0.5 tuned + 0.5 base & \WiseFiveLawRten & \WiseFiveTaxRten & \WiseFiveWikiRten & \WiseFiveBelRten \\
\midrule
BGE-M3 & \BgeLawRten & \BgeTaxRten & \BgeWikiRten & \BgeBelRten \\
BGE-M3, letters folded & \BgeFoldLawRten & \BgeFoldTaxRten & \BgeFoldWikiRten & \BgeFoldBelRten \\
\bottomrule
\end{tabular}
\caption{Recall@10 (percent) for the variants we tried. Folding maps the Tajik letters \tj{ҷ, ӣ, ӯ} (and capitals) to the Russian letters \tj{ч, и, у} before encoding.}
\label{tab:ablate}
\end{table}

\paragraph{bfloat16 training collapsed for these models.}
Our first test runs used bfloat16.
In two short test runs of a few dozen steps, one with the loss scale (inverse temperature) at 33.3 and one at 20, the value used later, the logged loss fell to exactly zero at step \BfZeroStep{} and stayed there (Figure~\ref{fig:loss}); the same settings in fp32 trained normally.
A loss of exactly zero in a contrastive objective is not a real value. The run log shows a non-finite gradient norm one step earlier, and we believe the training framework reports the resulting non-finite losses as zero; we did not confirm this further, but the weights of the bfloat16 run were no longer usable.
We did not see this with the two decoder-based embedders, which we trained with LoRA in bfloat16 without problems.
We therefore trained the XLM-RoBERTa-based embedders in fp32 and recommend checking the loss of the first steps before any long run.

\paragraph{Folding Tajik letters did not help.}
Because the tokenizer reads \UnkLetters{} as unknown, we tried replacing Tajik letters with their Russian look-alikes before encoding, at test time and without retraining: \tj{ҷ} to \tj{ч}, and also \tj{ӣ} to \tj{и} and \tj{ӯ} to \tj{у}, whose lower-case forms the tokenizer does know, with their capitals.
This made both XLM-RoBERTa-based models slightly worse (Table~\ref{tab:ablate}): the mean recall@10 over law, tax, Wikipedia and Belebele went from \BaseFourMean{} to \BaseFoldFourMean{} for E5 and from \BgeFourMean{} to \BgeFoldFourMean{} for BGE-M3.
The differences are small and we did not test them, but folding did not help in this setting, and we keep the original spelling.
We did not test why the folded text scores lower; one possibility is that folded words coincide with unrelated Russian words.

\paragraph{Weight interpolation and earlier checkpoints.}
Averaging the fine-tuned weights with the original ones \citep{wortsman2022wiseft} at 0.7 and 0.5 gave slightly lower law and tax recall, within the intervals, and left Belebele, Russian and English where they were (Table~\ref{tab:ablate}), so we saw no benefit.
Checkpoints at steps 300 and 400 score within noise of the final model on every set. We did not run a second epoch.

\section{Release}
\label{sec:release}

\paragraph{What we release.}
The model is on the Hugging Face Hub as \texttt{Tohirju/\ModelName}, in sentence-transformers format, together with the training log and the result files behind the scores in this report.
Access is given on request: the repository is public, and each request is approved by us.
The training pairs, the code and the full test questions are available on request; Table~\ref{tab:examples} shows examples of the test questions.
The customer-service pairs used in training are not released.

\paragraph{Licence.}
The starting model, multilingual-e5-large-instruct, is released under the MIT licence, and its notice is kept with our weights.
Our fine-tuned model is offered under the Saidzoda model licence: organisations with revenue need a paid licence, while startups and individuals without revenue may use it free of charge until they earn from it. Access requests go through \texttt{ai.saidzoda.com/access}.

\paragraph{How to use it.}
Load the model with sentence-transformers, put the original query instruction in front of each question, encode the passages without an instruction, and rank passages by cosine similarity.
The model was trained and tested with inputs of up to 512 tokens.

\paragraph{Intended use.}
The model is meant for finding Tajik passages that answer a Tajik question, for example as the first step of a question-answering system over laws or official documents.
The retrieved text should be checked before an answer reaches a user.
Its training included short customer-service requests, but we do not report a test on spoken or informal Tajik.

\section{Conclusion}

A single epoch of training on questions written by our own Tajik language model turned a 560M multilingual embedder into a Tajik search model that finds the right law or tax passage much more often than before, probably does slightly better on questions written by people, and keeps its Russian and English.
On our in-domain Tajik tests it is tied with models several times its size, which matters for anyone who has to run it on a single modest server; a larger model trained the same way is better on general reading questions.
The most useful next step is not more training but a set of questions written by Tajik speakers about real laws and tax rules, which will show how much of the gain holds outside our generated tests.

\section*{Limitations}

\paragraph{Same generator for training and test queries.}
The law, tax and Wikipedia questions in both training and test were written by the same model, on disjoint laws, documents and articles.
A retriever can learn the style of the generator's questions, so these scores are upper bounds.
Belebele is independent, but it consists of reading-comprehension questions about short news and travel texts, not the questions people ask about laws.
A set of questions written by Tajik speakers about our law and tax collections is the most important missing evaluation.

\paragraph{Statistics.}
The intervals are per-model bootstrap intervals; we did not store per-question ranks and so could not run paired tests.
The tax set has only \NTaxQ{} questions.
Each configuration was trained once, with one seed.
The starting model was chosen on the same law and tax test sets that we report, which slightly favours it in Table~\ref{tab:compare}.

\paragraph{Unequal comparisons.}
The two larger models were fine-tuned with LoRA on \LoraPct\% of the pairs, in bfloat16, while our model was fully fine-tuned on all pairs in fp32.
Their results are a lower bound on what those models can reach with our data. Their LoRA adapters had rank 32 and alpha 64 on all attention and feed-forward projections, with a learning rate of $10^{-4}$.
Latencies were measured on a shared GPU and are only indicative.

\paragraph{Scope.}
We tested retrieval only, not reranking or end-to-end question answering, and only text up to 512 tokens.
The customer-service training pairs come from a simulation and are not released, so that part of the training data cannot be reproduced by others.

\section*{Use of AI Assistants}

We used AI assistants (Anthropic Claude models) to write data-processing, training and analysis scripts, to search for and check references, and to draft parts of the text.
The training questions were generated by our own Tajik language model, as described in Section~\ref{sec:data}.
The authors designed the study, ran the training, checked every number against the recorded files, and take full responsibility for the content.
Every number in this report is generated by a script from logged files, and each reference was checked against its arXiv, ACL Anthology or DOI record.
AI assistants are not authors.

\section*{Acknowledgements}

We thank NVIDIA for access to the NVIDIA Inception programme, through which the training compute for this work was provided.

\bibliography{refs}
\bibliographystyle{acl_natbib}

\appendix
\section{Example Test Questions}
\begin{table*}[t]\centering\small
\begin{tabular}{lp{0.43\textwidth}p{0.38\textwidth}}
\toprule
Set & Question (Tajik) & English gloss \\
\midrule
Law & \tj{Мувофиқи моддаи} 43-\tj{и Кодекси андози Тоҷикистон, объектҳои муоинаи хронометражӣ кадомҳоянд}? & Under Article 43 of the Tax Code of Tajikistan, what are the objects of a timekeeping inspection? \\
Law & \tj{Тибқи Моддаи} 346-\tj{и Кодекси андози Ҷумҳурии Тоҷикистон, манбаи андоз чӣ тавр муайян карда мешавад}? & Under Article 346 of the Tax Code, how is the tax base determined? \\
Tax & \tj{Муқаррароти Кодекси одоби касбии мақомоти андоз ба кадом категорияи кормандон татбиқ мегардад}? & To which categories of staff do the rules of the tax authorities' code of professional ethics apply? \\
Tax & \tj{Корманди андоз бояд чӣ гуна амал кунад, то ба ошкоркунии сирри давлатӣ роҳ надиҳад}? & How must a tax officer act so as not to disclose state secrets? \\
Wikipedia & \tj{Дар замони Имперотурии Юан ҳокимияти имперотурӣ дар ҳар як вилоят аз ҷониби кадом намояндагон ташкил карда мешуд}? & In the Yuan Empire, which representatives exercised imperial authority in each province? \\
Wikipedia & \tj{Чаро дар интихоботи соли} 1924 \tj{Ҳизби Миллӣ ва ҳизби Меҳнатӣ пирӯз шуданд}? & Why did the National Party and the Labour Party win the 1924 election? \\
\bottomrule
\end{tabular}
\caption{Example test questions from the frozen law, tax and Wikipedia sets, exactly as written by the generator, without editing. The English glosses are ours. The law collection holds Tajik laws and codes, including the Tax Code; the tax collection holds tax-service orders, decrees and declaration forms. The full sets are available on request.}
\label{tab:examples}
\end{table*}

\end{document}